\documentclass[letterpaper]{article} 
\usepackage{aaai25}  
\usepackage{times}  
\usepackage{helvet}  
\usepackage{courier}  
\usepackage[hyphens]{url}  
\usepackage{graphicx} 
\usepackage{natbib}  
\usepackage{caption} 
\usepackage{algorithm}
\usepackage{algorithmic}
\usepackage{fancyhdr}       
\usepackage{graphicx}       
\usepackage{float}
\usepackage{subfig}
\graphicspath{{media/}}     
\usepackage{amsmath}
\usepackage{multirow}
\usepackage{makecell}

\usepackage{amsmath}
\usepackage{multirow}
\usepackage{makecell}

\usepackage{amssymb}
\usepackage{bbding}
\usepackage{pifont}
\usepackage{wasysym}
\usepackage{utfsym}
\usepackage{fontawesome}
\usepackage{appendix}
\usepackage{xcolor}
\definecolor{c1}{HTML}{0c8918}
\definecolor{c2}{HTML}{F4CAE4}

\usepackage{multirow}                 
\usepackage{multicol}                 
\usepackage{multirow}                
\usepackage{float}                    
\usepackage{makecell}                 
\usepackage{booktabs}                 

\usepackage{newfloat}
\usepackage{listings}
\DeclareCaptionStyle{ruled}{labelfont=normalfont,labelsep=colon,strut=off} 
\floatstyle{ruled}
\newfloat{listing}{tb}{lst}{}
\floatname{listing}{Listing}
\title{Set-CLIP: Exploring Aligned Semantic From \\ Low-Alignment Multimodal Data Through A Distribution View}
\author{
    Zijia Song\textsuperscript{\rm 1}, Zelin Zang\textsuperscript{\rm 2}, Yelin Wang\textsuperscript{\rm 3}, Guozheng Yang\textsuperscript{\rm 1}, \\Kaicheng Yu\textsuperscript{\rm 2}, Wanyu Chen\textsuperscript{\rm 1}, Miaoyu Wang\textsuperscript{\rm 3}, Stan Z. Li\textsuperscript{\rm 2\dag}
}
\affiliations{
    \textsuperscript{\rm 1}National University of Defense University\\
    \textsuperscript{\rm 2}Westlake University ~
    \textsuperscript{\rm 3}ShanghaiTech University\\
    \textsuperscript{\rm \dag}\{stan.zq.li@westlake.edu.cn\}\\

}

\usepackage{bibentry}

\begin{document}

\maketitle

\begin{abstract}
Multimodal fusion breaks through the boundaries between diverse modalities and has already achieved notable performances. However, in many specialized fields, it is struggling to obtain sufficient alignment data for training, which seriously limits the use of previously effective models. Therefore, semi-supervised learning approaches are attempted to facilitate multimodal alignment by learning from low-alignment data with fewer matched pairs, but traditional techniques like pseudo-labeling may run into troubles in the label-deficient scenarios. To tackle these challenges, we reframe semi-supervised multimodal alignment as a manifold matching issue and propose a new methodology based on CLIP, termed Set-CLIP. Specifically, by designing a novel semantic density distribution loss, we constrain the latent representation distribution with fine granularity and extract implicit semantic alignment from unpaired multimodal data, thereby reducing the reliance on numerous strictly matched pairs. Furthermore, we apply coarse-grained modality adaptation and unimodal self-supervised guidance to narrow the gaps between modality spaces and improve the stability of representation distributions. Extensive experiments conducted on a range of tasks in various fields, including protein analysis, remote sensing, and the general vision-language field, validate the efficacy of our proposed Set-CLIP method. Especially with no paired data for supervised training, Set-CLIP is still outstanding, which brings an improvement of $144.83\%$ over CLIP.
\end{abstract}

%

\section{Introduction}

As a pivotal foundation for numerous tasks\cite{stable_diffusion,anomalyclip}, multimodal learning has become the focus of many research\cite{ecor,splice}. By integrating information from diverse modalities such as texts, images and more, multimodal models can derive more comprehensive information to enhance the generalization of the learned representations\cite{survey_multimodal}. Meanwhile, such fusion enables networks to emulate human-like multiple perceptual capabilities and  address the inherent challenges like data scarcity, noise and ambiguity in various domains, from computer vision to healthcare\cite{medclip,zang2024dmt}. 

To better harness latent alignment information, previous studies have mainly concentrated on developing frameworks and pretraining objectives to enhance multimodal understanding. Result from the sufficient developments of images and texts, many studies have made great progress in the vision-language field\cite{clip,pali,vilt}.
Thereinto, CLIP employs a contrastive pretraining task on large-scale datasets and gets robust multimodal representation. Due to effective framework and general pretraining task, it has good portability and competitive performance compared with supervised methods, hence it becomes the baseline for various vision-language works\cite{flip}. Besides the aforesaid traditional field, in other intersecting domains, great breakthroughs have been also made by applying CLIP. EchoCLIP\cite{echoclip} improves the performance of cardiac imaging models by correlating ultrasound images with expert texts while ProtST\cite{xu2023protst} capture more protein function information by aligning protein sequences and textual property descriptions. Moreover, in the field of zero-shot video recognition, Open-VCLIP\cite{open-vclip} also shows excellent performance by leveraging the similar paradigm, which proves the powerful effects of CLIP.

\begin{figure*}[!t]
  \vspace{-0.3cm}
  \centering
  \subfloat[CLIP]{
  \centering
  \includegraphics[width=0.26\textwidth]{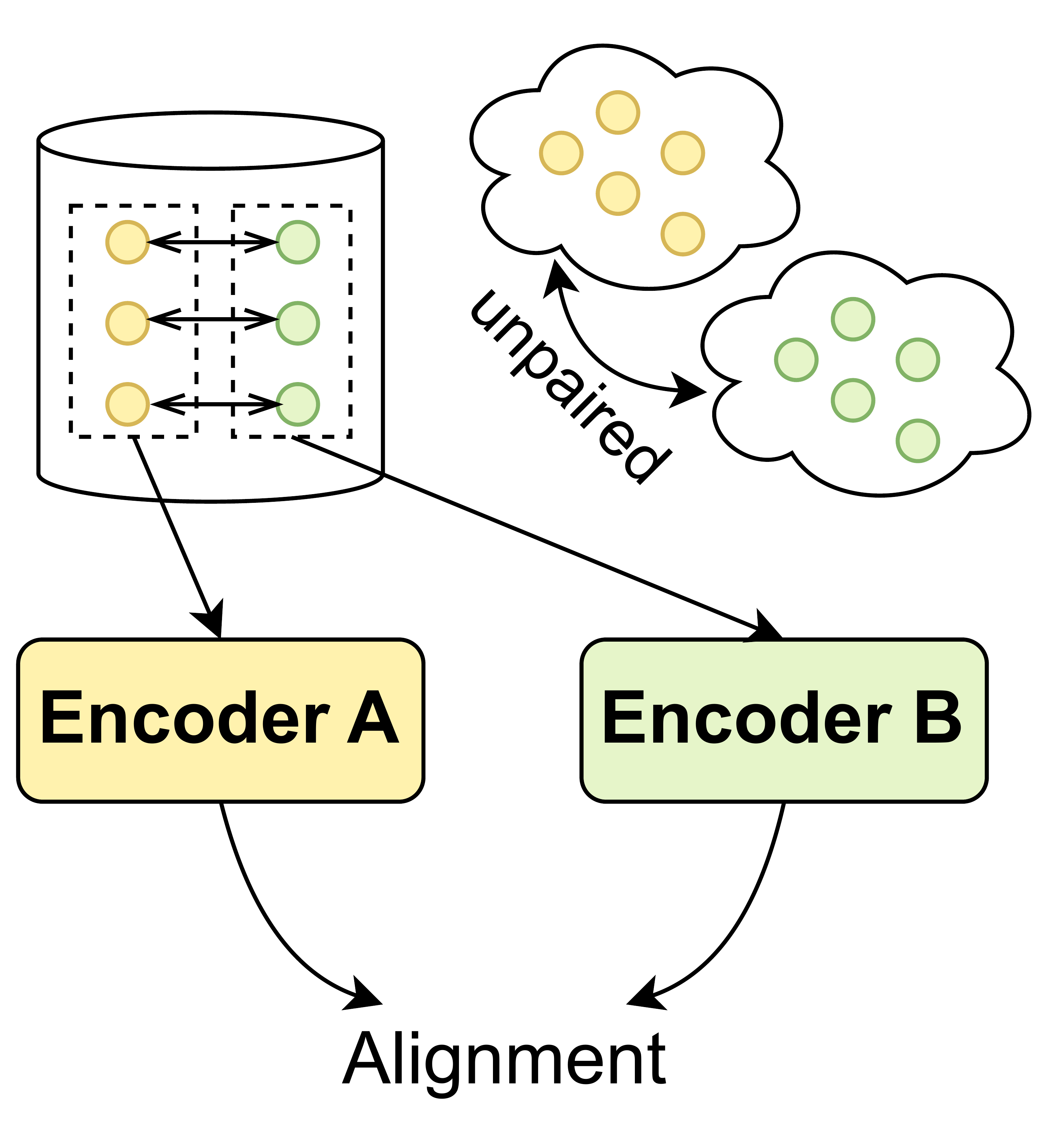}
  \label{clip}}
  \subfloat[S-CLIP]{
  \centering
  \includegraphics[width=0.3\textwidth]{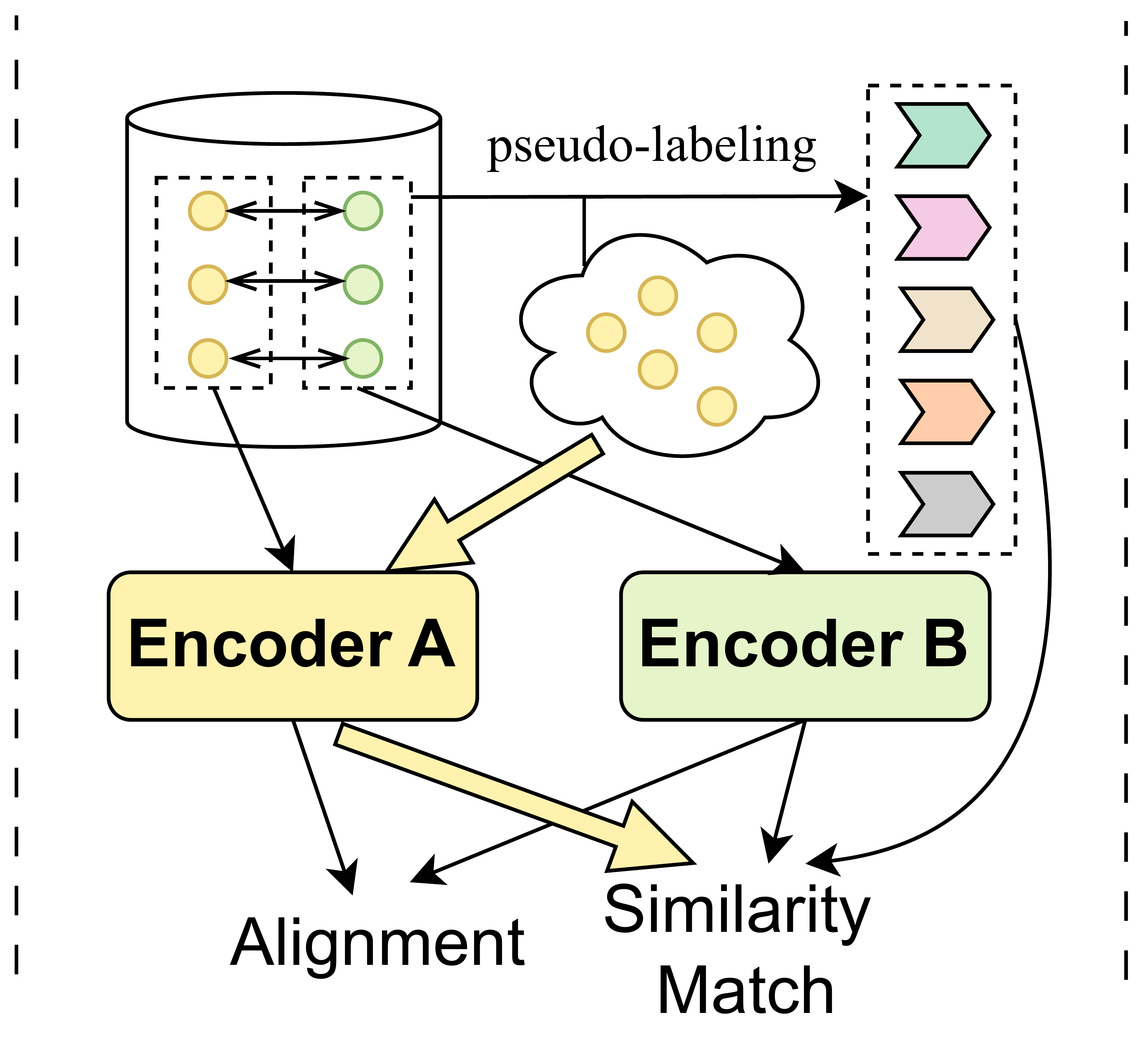}
  \label{s-clip}}
  \subfloat[Set-CLIP]{
  \centering
  \includegraphics[width=0.26\textwidth]{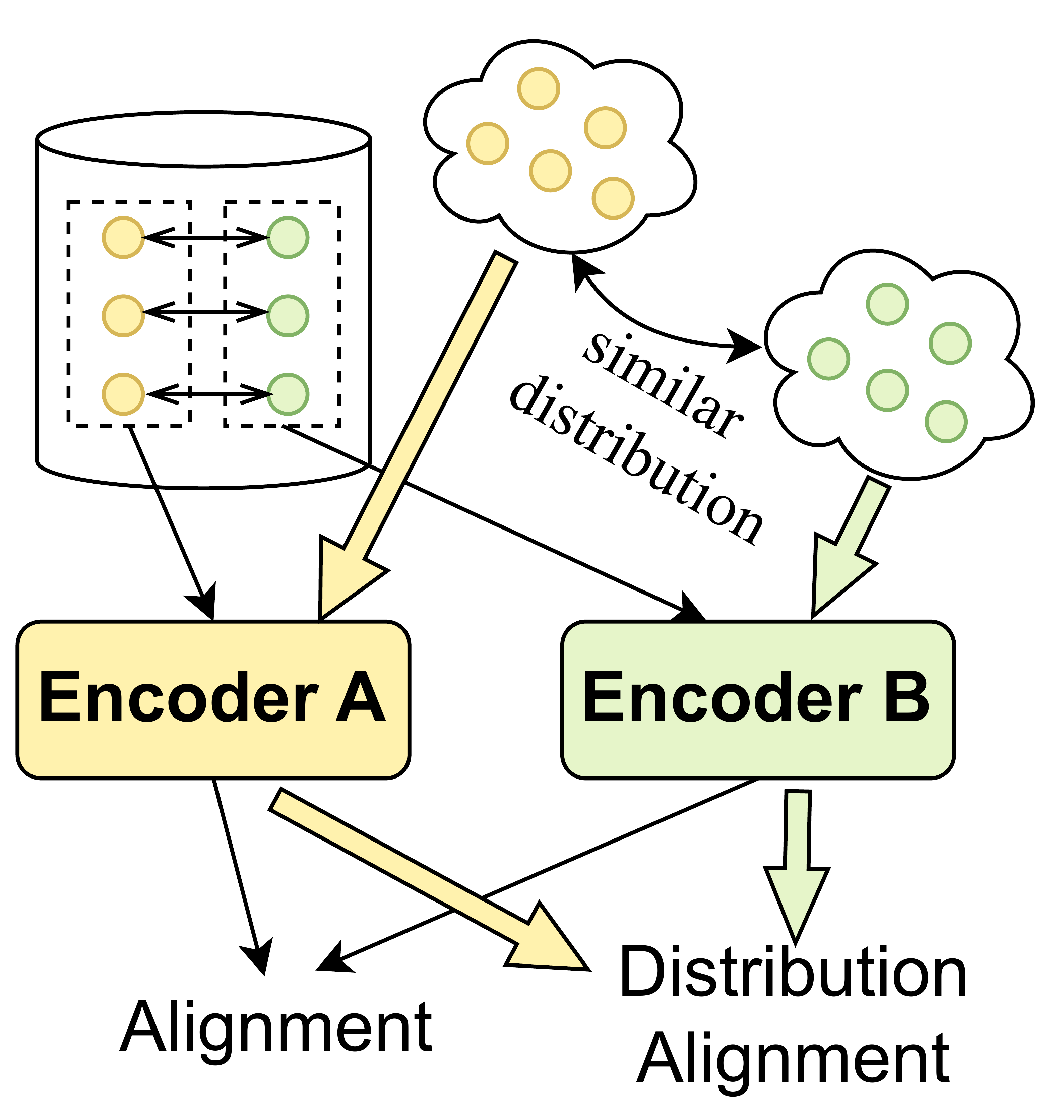}
  \label{set-clip}}
  \caption{Comparison among CLIP, S-CLIP and Set-CLIP on how to adopt unpaired multimodal data. (a) CLIP only uses the matched data for multimodal fusion while ignores the valuable information in unlabeled data. (b) S-CLIP attempts to improve the alignment performance by two pseudo-labeling losses but it limits itself to the language modality and heavily relies on the way how to measure similarity between samples. (c) Set-CLIP tries to explore more latent information from unmatched multimodal data by fine-grained distribution alignment, which is based on data themselves without much expert knowledge.}
  \label{fig1}
  \vspace{-0.4cm}
\end{figure*}

Nonetheless, there are still many specialized fields where it is usually difficult to obtain sufficient alignment data\cite{remote_limt_data} while traditional multimodal models like CLIP can only learn from matched pairs, which greatly limites the performance of previously elaborate models. In order to break above dilemma, several studies paid attention to these specialized fields and attempted to learn from low-alignment data with fewer matched pairs for pretraining\cite{s-clip_realted_work_2,s-clip_realted_work_4}. The main idea is to modify the loss function of CLIP and apply semi-supervised learning method\cite{survey_semi_supervised_2} to explore the latent alignment information from unlabeled data. Recently, a new research improves the original CLIP and proposes S-CLIP\cite{s-clip} which introduces two novel pseudo-labeling losses for unlabeled images and achieve state-of-the-art in various specialized semi-supervised vision-language fields. 

However, pseudo-labeling methods may only be limited to the fields with class information and have difficulties scaling to other specialized multimodal domains where pseudo-labels are struggling to obtain. Meanwhile, the knowledge of generating pseudo-labels only relies on the insufficient labeled data, which leads to narrow ken and may loss much potential alignment information. In addition, the quality of pseudo-label has a great impact on the final performance so the learning process is unstable and even negative\cite{pseudo_limit}. In order to solve these problems, it is necessary to design new semi-supervised methods for multimodalities, which can capture latent alignment information in unpaired data and be well extended to various multimodal domains. 

Therefore, we propose a novel semi-supervised learning method for multimodal alignment based on CLIP, named as Set-CLIP. We believe that ultimate representation is composed of modality, structure as well as semantic and the key to multimodal alignment is to capture the same semantic representation while ignoring the other two parts. On the premise of two aligned modal data with the same semantic distribution, we design a new pretraining task based on manifold matching and a novel loss called semantic density distribution(SDD) to better concentrate on the implicit alignment among vast unpaired multimodal data. Moreover, we introduce multi-kernel maximum mean discrepancy(MK-MMD) to eliminate the gap between modality representations while self-supervised contrastive loss is used to prevent mode collapse and enhance the robustness of semantic representation. At the same time, we apply contrastive loss from CLIP on the matched multimodal pairs to keep the correct learning direction. Set-CLIP tries to explore alignment relationship in latent space and it can be extended to various multimodal domains due to task irrelevance. Through end-to-end learning, the mutual constraints between losses prevent negative optimization and implicitly expand the knowledge range. Our approach can be transferred to different multimodal frameworks\cite{blip,glip} and the comparison of Set-CLIP with other strategies is shown in Figure~\ref{fig1}.

In short, our contributions are summarized as follows:
(1) We contribute a groundbreaking perspective for the semi-supervised multimodal alignment problem by reframing it as a manifold matching problem, which brings a new pathway to exploit the implicit alignment information in the rich, yet largely unmatched multimodal data. 
(2) We design a novel semantic density distribution loss with fine-grained constrain and it can be applied in various specialized fields as well as different multimodal frameworks. We introduce other objectives based on theoretical analysis about the components of representation and propose Set-CLIP to realize multimodal alignment with less supervised pairs. Moreover, our method can be applied to other domains with two-stream networks\cite{two-stream}, such as knowledge distillation\cite{self-distillation}, self-supervised learning\cite{moco} and domain adaptation.
(3) We conduct extensive experiments in various fields and prove the advantages of Set-CLIP. Moreover, We also explain the effects of key modules and provide a feasible usage paradigm for the specialized fields with limited supervised pairs.



\section{Related Works}
\textbf{Multimodal alignment.}
Multimodality enhances understanding and decision-making by integrating information from multiple sensorymodalities\cite{martin2022multimodality}. Thereinto, ALBEF \cite{li2021align} aligns visual and language representations, using momentum distillation to improve multimodal embeddings. FLAVA \cite{singh2022flava} enhances multitask and cross-modal learning by jointly pretraining text and images. ALIGN \cite{jia2021scaling} jointly trains language and image encoders, significantly enhancing performance across various vision and text benchmarks. In recent years, researches around CLIP has further optimized computational efficiency and model representation capabilities.
For instance, FLIP\cite{flip} brings lower computation and faster training times by randomly removing a large number of image patches during training process while SoftCLIP\cite{softclip} applies fine-grained interior self-similarity as a softening target to alleviate the strict mutual exclusion problem. Moreover,  latent diffusion models generates reliable text embeddings as condition by using pretrained text encoder of CLIP and CLIPSelf \cite{clipself} enhances region-level representation through self-distillation from CLIP's image encoder, which proves the powerful effects of CLIP.

\textbf{Semi-supervised learning.}
Semi-supervised learning \cite{van2020survey} uses both labeled and unlabeled data to improve training process, encompassing strategies like pseudo-labeling \cite{cascante2021curriculum}, where models self-label their training data, and self-supervised learning \cite{krishnan2022self,liu2022graph}, which explores the values in data itself. vONTSS \cite{xu2023vontss} utilizes the von Mises-Fisher distribution and optimal transport for semi-supervised neural topic modeling to improve topic extraction in text datasets. SSGD \cite{zhou2023semi} proposes a new semi-supervised domain generalization method that enhances model robustness under domain shifts through stochastic modeling and style augmentation. SS-ORL \cite{zheng2023semi} employs a semi-supervised offline reinforcement learning approach, improving learning outcomes by utilizing unlabeled trajectories and limited complete action data. Semi-supervised learning ensures performance with fewer samples, but in the specialized domains, how to achieve complementarity and integration across different modalities with limited paired data remains an issue that needs attention and resolution\cite{ssml}.


\section{Method}
\subsection{Problem Description and Assumption}
Different from the general vision-language field, there could be only limited available matched pairs between specific associated modalities while it is relatively simple to get a large amount of unimodal data with similar semantic distribution. Therefore, we propose Set-CLIP, which uses massive unmatched data as well as limited matched pairs to realize more generalized alignment through semi-supervised learning. Formally, for any two modalities $\bf A$ and $\bf B$, we employ a small number of matched pairs $\left\{a_i, b_i\right\}_{i=1}^N$ and a large number of unmatched data $\left\{a_j \in \bf A\right\}_{j=1}^{M_1}$ as well as $\left\{b_j \in \bf B\right\}_{j=1}^{M_2}$ to train our model. Through sampling respectively from two unpaired sets, we acquire $\left\{a_j \in \bf A\right\}_{j=1}^M$ and $\left\{b_j \in \bf B\right\}_{j=1}^M$ as unsupervised training data which are only considered to have similar semantic distribution rather than strict one-to-one matching. Based on a natural assumption, models can be trained on these adequate unmatched multimodal data.

\textbf{Assumption 1 (Semantic Distribution Similarity Assumption, SDSA).} We suppose that the latent embedding is a combination representation of modality, structure as well as semantic and more detailed analysis will be displayed in Appendix A. The goal of multimodal alignment is to find the same semantic representation and get rid of the interference from the other two representations. If the overall semantic distributions of $\left\{a_j \in \bf A\right\}_{j=1}^M$ and $\left\{b_j \in \bf B\right\}_{j=1}^M$ are similar, we can find a embedding space $\bf{S} \subseteq \bf{R} ^K$ where $u_j$ and $v_j$ are the embedding representations respectively from $\bf{A}$ and $\bf{B}$. When the density distributions of $\left\{u_j \in \bf S\right\}_{j=1}^M$ as $\bf U$ and $\left\{v_j \in \bf S\right\}_{j=1}^M$ as $\bf V$ are similar, this space $\bf S$ is the semantic embedding space of $\bf A$ and $\bf B$. Consequently, when datasets from two modalities have the similar semantic distribution and their volumes are large enough, we can find the aligned semantic space by narrowing the gap between the density distribution from two modalities rather than strict matching relationship or pseudo-labeling method. Through the above assumption, we can explore the value from unpaired data and the semi-supervised multimodal alignment can be transformed into a manifold matching problem.

\subsection{The Framework of Set-CLIP}
\begin{figure*}
  \centering
  \includegraphics[width=0.9\textwidth]{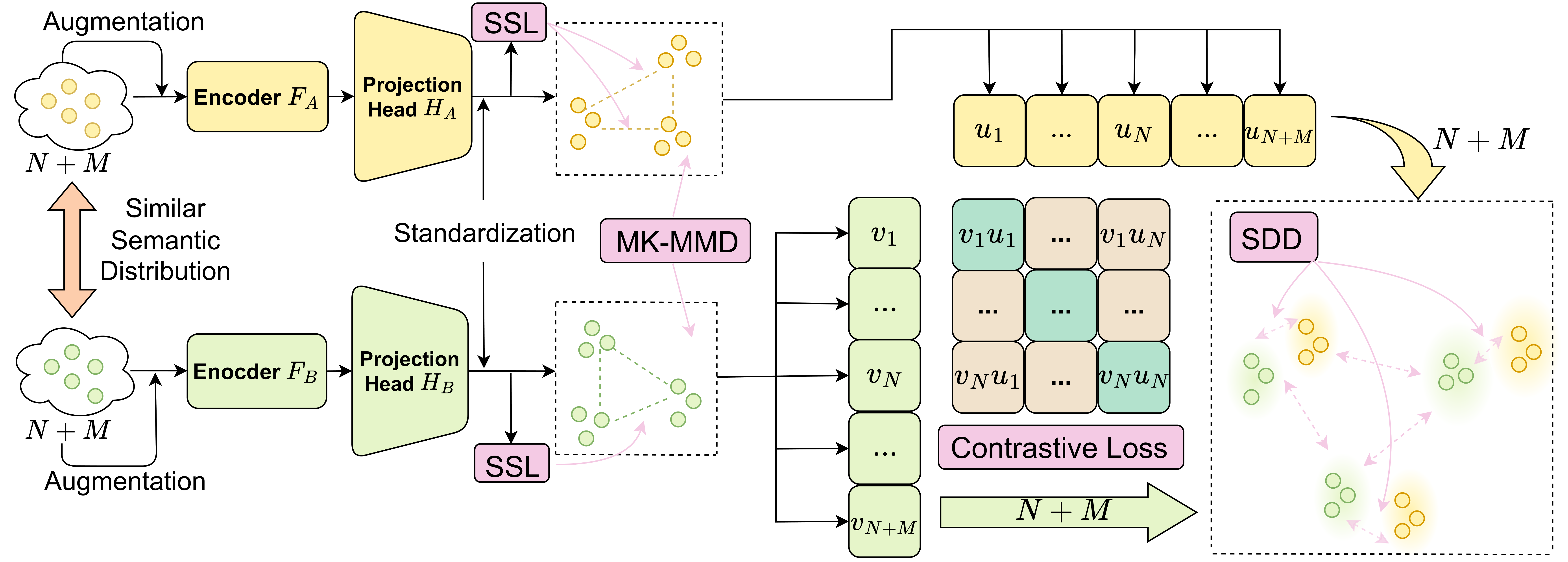}
  \caption{The overall framework of Set-CLIP. Here, $N$ is the number of matched pairs, while $M$ denotes the unlabeled scale. \textcolor{c2}{Pink} indicates the loss objectives. Thereinto, MK-MMD is used to narrow the gap between the distribution spaces of different modalities, and SDD operates to reduce the differences between latent distributions in a fine-grained way. Applying SSL can enhance the robustness of representation, and the contrastive loss in CLIP ensures the proper optimization direction.}
  \label{fig2}
  \vspace{-0.3cm}
\end{figure*}

Figure~\ref{fig2} introduces the conceptual overview of Set-CLIP. Due to the convenience and efficiency of CLIP, our proposed method follows to design the two-stream network. Each stream includes an encoder network $F_i(\cdot),i\in\{ \bf A,\bf B\}$ and a projection head network $H_i(\cdot),i\in\{ \bf A,\bf B\}$, which are applied to map the data from original space into embedding space. The network from different streams adopt different backbone and is trained from scratch. We introduce MK-MMD as well as self-supervised contrastive loss(SSL) and design a novel semantic density distribution loss(SDD) to learn potential alignment in large amounts of unpaired data. Through contrastive metrix, we apply contrastive loss(CL) on limited supervised pairs to guarantee proper optimization. The multimodal batch with the size of $B$ is composed of $\lfloor \frac{N}{M+N}B \rfloor$ paired data from $\left\{a_i, b_i\right\}_{i=1}^N$ while the rest data is sampling from two unsupervised training datasets. A detailed description of loss functions will be shown below.

\subsection{Objective Loss}
\textbf{Coarse-Grained Modality Adaptation:}
There may be large discrepancy between the embedding distributions from different modalities. While the latent representations are supposed to have the similar distribution space when they become aligned, which can be treated as a domain adaptation problem. Thereinto, MK-MMD\cite{dan} is used to measure the gap between two probability distributions $P$ and $Q$ while the core idea of this method is that samples $\{p_1, p_2, \ldots, p_m\}$ as well as $\{q_1, q_2, \ldots, q_n\}$ drawn from $P$ and $Q$ should keep similar statistical properties if the two distributions are the same. Specifically, MK-MMD maps the data from original space to Reproducing Kernel Hilbert Spac(RKHS) by kernel functions\cite{kernel_function} and we can compare the difference between distributions in this space. Through linear combination of multiple kernel functions, we could get a more robust mapping function to RKHS where we can easily distinguish two distributions even though they are similar in original space. The formula is shown as follows:
\begin{equation}
\mathcal{L}_\mathrm{MK-MMD}=\left\|\frac{1}{B} \sum_{i=1}^B\phi\left(u_i\right)-\frac{1}{B} \sum_{j=1}^B \phi\left(v_j\right)\right\|_{\mathcal{H}_k}^2
\end{equation}
where $B$ is batch size while $u_i$ and $v_j$ are latent representations from two modalities. $\mathcal{H}_k$ is RKHS induced by kernel function $k$ and $\phi(\cdot)$ is implicit function used to map the original space data to $\mathcal{H}_k$. For multi-kernel cases, kernel function $k$ is a linear combination of $d$ basic kernel functions $\{k_1, k_2, \ldots, k_d\}$ and the format is $k=\sum_{i=1}^d\beta_i k_i$. learnable kernel weight $\beta_i$ is  obtained through optimization to effectively represent differences between distributions. In our method, $d$ equals to $2$ while we choose Gaussian Kernel and Polynomial Kernel as basic kernel function.

\textbf{Fine-grained Semantic Distribution Alignment:}
Since MK-MMD pays attaention to the whole distribution rather than sample level so it is imprecise and can only achieve macro alignment which is not enough for representation alignment. Consequently, we propose a novel objective named as semantic density distribution loss(SDD) to explore more fine-grained information from unpaired data and realize more refined alignment. SDD is inspired from the perspective of probability density distribution estimation, hence it could keep an eye on specific sample representation while take the whole semantic distribution alignment into consideration at the same time. The formula is shown as follows:
\begin{equation}
\mathcal{L}_\mathrm{SDD}=\frac{1}{2}[\Gamma(U,V)+\Gamma(V,U)]
\label{eight}
\end{equation}
where $\mathcal{L}_\mathrm{SDD}$ works on the embedding space to measure the difference between two representation distributions more accurately in a symmetrical way and the models are trained to minimize the loss value to realize latent semantic alignment. $U$ and $V$ denotes embedding distributions and the format of $\Gamma(\cdot,\cdot)$ is shown as follows:
\begin{equation}
\Gamma(T,R)=\sum_{i=1}^B \{\frac{\kappa(t_i,T)}{\sum_{j=1}^B \kappa(t_j,T)}\log\frac{\kappa(t_i,T)/\sum_{j=1}^B \kappa(t_j,T)}{\kappa(t_i,R)/\sum_{j=1}^B \kappa(t_j,R)}\}
\label{four}
\end{equation}
here for generality and convenience, we define three intermediate variables, $T=\{t_i\}_{i=1}^B$ while $R=\{r_j\}_{j=1}^B$ are sets composed of latent represenations and $x$ denotes the latent representation of a sample. $B$ is the size of batch which is a combination of matched pairs and unmatched data and Kullback-Leibler divergence is introduced to measure the dis-similarity between the density values of a specific sample from two distributions. The format of $\kappa(\cdot,\cdot)$ is displayed in the following formula.
\begin{equation}
\kappa(x,T)=\frac{\sum_{i=1}^B  \exp \left(-\frac{\| x-t_i \|^2}{ b^2\sigma(T)}\right)}{2Bb^2 \pi }
\label{five}
\end{equation}
here we apply exponential function as probability density function and $b$ denotes bandwidth used to control the smoothness. $\sigma(\cdot)$ denotes the variance of distribution and the format is shown as follows.
\begin{equation}
\sigma(T)=\frac{1}{B-1}\sum_{i=1}^B\left\|t_i-\frac{\sum_{j=1}^Bt_i}{B}\right\|^2  
\label{six}
\end{equation}
where $t_i$ is the sample from set $T$ and we apply sample variance with Bessel's Correction. $\sigma(\cdot)$ can lead model to focus on narrowing the gap between semantic distributions while avoid close cluster. By employing $\mathcal{L}_\mathrm{SDD}$, semantic aligned data from different modalities will get similar density distribution in the latent space during training. Meanwhile, the time complexity of $\mathcal{L}_\mathrm{SDD}$ is $\mathcal{O}(B^2)$ which is the same as $\mathcal{L}_\mathrm{CL}$. More details of SDD will show in Appendix B.




\textbf{Supervised Alignment Guidance:}
Based on problem description, there are limited matched pairs $\left\{a_i, b_i\right\}_{i=1}^N$ and a large number of unmatched data $\left\{a_j, b_j\right\}_{j=1}^M$. Due to the lack of sufficient data, we are supposed to learn generalized representations through unsupervised data and take advantage of explicit alignment relationship as ground-truth to achieve precise alignment. We apply contrastive loss $\mathcal{L}_{\mathrm{CL}}$ in CLIP, which is to maximize the representation similarity between matched pairs while minimize the similarity between negative pairs. The format of this loss is shown as follows:
\begin{equation}
\begin{split}
\mathcal{L}_{\mathrm{CL}}=-\frac{1}{2 n} \sum_{i=1}^n( &\log \frac{\exp \left(\mathcal{S}(u_i,v_i) / \tau\right)}{\sum_{j=1}^n \exp \left(\mathcal{S}(u_i,v_j) / \tau\right)}\\
+ &\log \frac{\exp \left(\mathcal{S}(v_i,u_i) / \tau\right)}{\sum_{j=1}^n \exp \left(\mathcal{S}(v_i,u_j) / \tau\right)} )
\end{split}
\end{equation}
where $n$ denotes the paired size in a batch and is generally  $\lfloor \frac{N}{M+N}B \rfloor$. $u_i$ and $v_i$ are representations in latent space respectively from two modalities. $\tau$ is a learnable temperature parameter and $\mathcal{S}(\cdot,\cdot)$ denotes cosine similarity. We expect to apply supervised as well as unsupervised data in every batch to jointly train the model due to the reason that a mass of unsupervised data can bring richer alignment information while matched pairs could lead to more accurate learning.

\textbf{Self-supervised Distribution Stability:}
Rely on self-supervised contrastive loss(SSL)\cite{simclr,simcse}, we can adequately find out implicit information from single modality and get robust feature representation. In the field of multimodal alignment with limit matched pairs, we find that it is essential to apply this objective because it can pull away the representations of different samples in the latent space with incomplete alignment guidance. In other words, if SSL is not employed, the data without alignment constraint may gather into a tight cluster. To be specific, we apply augmentation to generate positive pairs and the format of $\mathcal{L}_\mathrm{SSL}$ is displayed as follows:
\begin{equation}
\mathcal{L}_\mathrm{SSL} = -\frac{1}{B} \sum_{i=1}^{B} \log \frac{\exp(\mathcal{S}(z_i, z_{i}^{+}) / \tau)}{\sum_{j=1}^{B} \exp(\mathcal{S}(z_i, z_j) / \tau)}
\end{equation}
where $z$ is latent embedding and $z_{i}^{+}$ denotes the representation of corresponding positive sample. In our method, each modality is supposed to apply this loss while $\mathcal{S}(\cdot,\cdot)$ is calculated through cosine similarity. We denote $u_i$ and $v_i$ as practical latent representation respectively from different two modalities and the corresponding objectives are named as $\mathcal{L}_\mathrm{SSL-U}$ as well as $\mathcal{L}_\mathrm{SSL-V}$. According to the above constraint, we propose new loss named as $\mathcal{L}_{GC}$ and the formula is shown as follows:
\begin{equation}
    \mathcal{L}_{GC}=\mathcal{L}_{CL}+\mu\mathcal{L}_{SSL-U}+ \\
    \mu\mathcal{L}_{SSL-V}
\end{equation}
where $\mu$ is a hyperparameter and $\mathcal{L}_{GC}$ is used to guide the training process with accurate supervised alignment information rather than semantic distribution similarity, which is necessary for avoiding negative optimization. Meanwhile, if the data from different modalities can achieve augmentation according to the common semantics rather than the pattern in the single modality, the performance of related method may realize further growth\cite{plato}.

\textbf{The Overall Pretraining Objective:}
Our method aims to adopt matched pairs as well as unsupervised data in a batch at the same time. In this way, during the pretraining process, we can utilize comprehensive unsupervised data as well as the alignment constraint from matched pairs to realize robust and stable optimization process. Moreover, through semantic distribution alignment, the knowledge learned from unsupervised data and matched pairs can potentially interact with each other which could enlarge the range of knowledge. For overall pretraining objective, we seek to minimize the loss functions of all pretraining tasks simultaneously:
\begin{equation}
\mathop{\min}_{\theta} ~\alpha \mathcal{L}_\mathrm{GC}+\delta \mathcal{L}_\mathrm{MK-MMD}+\eta \mathcal{L}_\mathrm{SDD}
\end{equation}
where $\theta$ denotes all learnable parameters in encoder and projection head networks. $\alpha$, $\delta$ and $\eta$ are hyperparameters used to control the impacts of different pretraining tasks.


\section{Experiments}
In order to evaluate the effectiveness of proposed method, we conduct extensive experiments in various fields, including protein representation, remote sensing as well as general vision-language field. In addition, we design sufficient ablation experiments to analyze the roles of key modules.

\subsection{Quantitative Analysis About Sampling Size}
As mentioned above, there exists implicit alignment information between different modalities with similar semantic distribution even if there is no definite matched pairs. Therefore, if we can acquire unimodal batches which reflect the real distribution of original data by stochastic sampling in each modality, it is derivable that each batch from different modalities also keeps similar semantic distribution and can be used for subsequent training process. Obviously, sampling size significantly influence the ability whether batches are on behalf of original distributions. Hence We attempt to quantitatively analyze the ability of different scales of sampling size for representing the original distribution, which can guide to choose the proper size. By applying soft Parzen-window method, we can calculate the representing confidence of sample batch with given size.  Through experimental verification, it could be concluded that sample batches will be able to represent original complicated distribution effectively when sampling size is over $64$. Furthermore, if different modal data is from the same semantic distribution, the batches with sampling size over $64$ will also keep the similar semantic distribution. More Detailed process of method and relevant analysis will be displayed in Appendix E.

\begin{table}[t]
\centering
\scalebox{0.89}{
\begin{tabular}{cccccc}
\toprule
\multirow{2}{*}{Method} & \multicolumn{3}{c}{Gene Ontology}                & \multirow{2}{*}{EC} & \multirow{2}{*}{Average}             \\ \cline{2-4}
                        & BP             & MF             & CC             &                                                                                                             \\ \midrule
ResNet                    & 0.280          & 0.405          & 0.304          & 0.605                                                                        & 0.399                                                                     \\
                        ProtBert                  & 0.279          & 0.456          & 0.408          & 0.838                                                                        & 0.495                                                                     \\
                        OntoProtein               & 0.436          & 0.631          & 0.441          & 0.841
                                                & 0.587                \\
                        ESM-1b                    & 0.452          & 0.659          & 0.477          & 0.869                                                                        & 0.614                                                                     \\
                        ESM-2             & \textbf{0.472}          & 0.662          & 0.472          & 0.874                                                                        & 0.620                                                                     \\ \hline
GraphQA             & 0.308       & 0.329          & 0.413          & 0.509
                                                & 0.389                \\
                        GVP                 & 0.326       & 0.426          & 0.420          & 0.489
                                                & 0.415                \\
                        DeepFRI             & 0.399       & 0.465          & 0.460          & 0.631
                                                & 0.489                \\
                        GearNet                 & 0.356          & 0.503          & 0.414          & 0.730
                                                & 0.501                \\
                        New IEConv          & 0.374       & 0.544          & 0.444          & 0.735
                                                & 0.524                \\
                        GearNet-Edge        & 0.403       & 0.580          & 0.450          & 0.810
                                                & 0.561                \\
                        CDConv                  & 0.453          & 0.654          & 0.479          & 0.820                                                                        & 0.602                 \\ \cline{1-6} 
                        CLIP(1/2 CATH)          & 0.456              & 0.661              & 0.485              & 0.881        
                                    & 0.621                     \\
                        Set-CLIP(Ours)          & 0.459 & \textbf{0.667} & \underline{0.491} & \underline{0.884}                                                               & \multicolumn{1}{c}{\underline{0.625}}  \\ 
                        CLIP(CATH)              & \underline{0.463}          & \underline{0.665}              & \textbf{0.493}              & \textbf{0.885}         
                                    & \textbf{0.627}                     \\\bottomrule
\end{tabular}}
\caption{Benchmark results on protein representation field. \textbf{Bold} denotes the best results while \underline{underline} indicates the second best value. Two-stream networks like CLIP(CATH) outperform in most tasks and Set-CLIP can effectively reduce the gap between CLIP(CATH) with only half paired data of CATH for supervised alignment.}
\label{table2}
\vspace{-0.3cm}
\end{table}

\begin{table*}[t]
\centering
\scalebox{1}{
\begin{tabular}{clllll}
\toprule
Method                           & RSICD-CLS & UCM-CLS & WHU-RS19 & RSSCN7 & AID  \\ \midrule
CLIP(original)                      & 45.3      & 50.5    & 65.5     & 58.9   & 47.8 \\ \hline
CLIP(fine-tune)                     & 58.3{\scriptsize $\pm$0.3}      & 63.5{\scriptsize $\pm$3.4}    & 76.5{\scriptsize $\pm$3.2}     & 61.9{\scriptsize $\pm$1.2}   & 63.1{\scriptsize $\pm$1.3} \\ \hline
Hard-PL           & 56.6{\scriptsize $\pm$3.5}      & 61.6{\scriptsize $\pm$2.2}    & 78.1{\scriptsize $\pm$2.5}     & 63.9{\scriptsize $\pm$2.1}   & 63.2{\scriptsize $\pm$2.6} \\
Soft-PL                               & 62.5{\scriptsize $\pm$0.8}      & 65.7{\scriptsize $\pm$2.7}    & 83.7{\scriptsize $\pm$2.7}     & 65.7{\scriptsize $\pm$0.6}   & 68.0{\scriptsize $\pm$0.7} \\
S-CLIP                                & 66.9{\scriptsize $\pm$1.7}      & 66.7{\scriptsize $\pm$1.6}    & 86.9{\scriptsize $\pm$2.0}     & \textbf{66.2}{\scriptsize $\pm$1.1}   & 73.0{\scriptsize $\pm$0.3} \\  
Set-CLIP(ours)        & \textbf{69.2}{\scriptsize $\pm$0.8}      & \textbf{67.5}{\scriptsize $\pm$1.1}    & \textbf{89.0}{\scriptsize $\pm$1.6}     & \textbf{66.2}{\scriptsize $\pm$0.9}   & \textbf{76.2}{\scriptsize $\pm$0.9} \\ \bottomrule
\end{tabular}}
\caption{Benchmark results on remote sensing field. \textbf{Bold} is the best average results and Set-CLIP improves the performance at most datasets through learning from unsupervised multimodal data.}
\label{table3}
\vspace{-0.4cm}
\end{table*}

\subsection{Evaluation On Single Protein Function Prediction}
\textbf{Overview of tasks and training setup:}
To examine the efficacy of Set-CLIP in non-vision-language multimodal domains with insufficient alignment data, we conduct experiments in the protein representation field. Proteins can be defined using a multi-level structure and most previous works take aligned sequence and structure as input for single-stream network to capture the invariance features\cite{ieconv_atom}. Due to limited aligned data, these intricately designed models struggle into trouble. Following \cite{zheng2023lightweight}, we consider sequence and structure as two modalities and apply Set-CLIP to realize multimodal fusion by pulling semantic distributions closer from extensive unsupervised data. Structure encoder is designed based on CDConv\cite{cdconv} while ESM-2\cite{esm2} is selected as sequence encoder. We adopt CATH $4.2$ dataset for pretraining and this process lasts $100$ epochs. According to the same settings in \cite{cdconv}, we evaluate the proposed method on the following four tasks: protein fold classification, enzyme reaction classification, gene ontology (GO) term prediction and enzyme commission (EC) number prediction\cite{go_and_ec}. More details of this experiment is shown in Appendix F and D.



\textbf{Results:}
The performance of downstream tasks are shown in Table~\ref{table2} while results of previous approaches are from \cite{cdconv,gearnet,xu2023protst}. Thereinto, CLIP(1/2 CATH) is a two-stream network with 50\% CATH data for pretraining while CLIP(CATH) is pretrained on the whole CATH datasets. Set-CLIP(Ours) adopts 50\% CATH data as supervised pairs while the rest are considered as unlabeled data. Moreover, we add an Average item to evaluate the overall performance. We first verify the effect of two-stream network compared to single-stream model and the results are displayed at the last line in Table~\ref{table2}. We can find that CLIP achieve better results at most downstream tasks and show superiority especially in EC number prediction. Further, it is obvious that Set-CLIP dramatically narrow the overall gap between CLIP(CATH) and the performance is even better at a few downstream tasks. This may be due to the fact that Set-CLIP can explore implicit alignment through the fine-grained semantic distribution constraint of SDD while CLIP only focuses on local representation match which may loss some global distribution information.  



\subsection{Evaluation On Remote Sensing Datasets}
\textbf{Overview of tasks and training setup:}
The models in remote sensing field can acquire comprehensive knowledge by jointly learning satellite images and corresponding captions. However, the training datasets are usually composed of web-crawled data and annotating captions may also need various expert knowledge, which can be expensive and time-consuming. So it is essential to evaluate the performance of Set-CLIP on limited matched pairs which is hard to tackle by traditional methods. Following \cite{s-clip}, Set-CLIP is pretrained on the union of RSICD, UCM and Sydney with zero-shot classification and image-text retrieval as downstream tasks. ResNet\cite{resnet} and transformer\cite{transformer} are chosen as encoders and Set-CLIP is pretrained for $25$ epochs. We subsample $10\%$ of image-text pairs for supervised learning while the remaining data is served unlabeled but conform to the same semantic distribution. Similarly, Top-1 classification accuracy is used to evaluate the performance on zero-shot classification while recall is applied for image-text retrieval tasks. More Details will be presented in the Appendix C and D.



\begin{figure*}[t]
\begin{minipage}{.68\textwidth}
\centering
\vspace{-0.2cm}
\scalebox{0.9}{
\begin{tabular}{ccrrrrrrrr}
\toprule
\multirow{4}{*}{Model}   &\multirow{4}{*}{Method}   & \multicolumn{4}{c}{\multirow{1.25}{*}{Flickr-8k}}     & \multicolumn{4}{c}{\multirow{1.25}{*}{Mini COCO}}     \\ \cmidrule(lr){3-6} \cmidrule(lr){7-10} 
\multicolumn{2}{c}{}                           & \multicolumn{2}{c}{Image $\rightarrow$ Text} & \multicolumn{2}{c}{Text $\rightarrow$ Image} & \multicolumn{2}{c}{Image $\rightarrow$ Text} & \multicolumn{2}{c}{Text $\rightarrow$ Image} \\ \cmidrule(lr){3-4} \cmidrule(lr){5-6} \cmidrule(lr){7-8} \cmidrule(lr){9-10}
\multicolumn{2}{c}{}                           & top1   & top5  & top1  & top5   & top1   & top5  & top1   & top5  \\ \midrule
\multirow{3}{*}{ResNet} & CLIP(1/3)       & 11.6               & 35.5               & 10.9               & 34.2               & 7.9                 
                         & 30.6                  & 8.2                   & 30.3     \\
                        & CLIP(1)         & \color{c1}+9.5   & \color{c1}+17.4  & \color{c1}+8.9   & \color{c1}+18.1  & \color{c1}+6.5      & \color{c1}+16.3     & \color{c1}+6.0      & \color{c1}+16.2      \\ 
                        & Set-CLIP     & \color{c1}+1.7   & \color{c1}+5.4  & \color{c1}+1.2    & \color{c1}+5.2   & \color{c1}+3.8      & \color{c1}+8.5      & \color{c1}+3.1      & \color{c1}+8.5      \\ \hline
\multirow{3}{*}{VIT}    & CLIP(1/3)       & 14.7              & 42.7             & 14.5              & 41.7             & 10.5                & 37.1                                              & 9.7                & 36.5       \\
                        & CLIP(1)         & \color{c1}+10.9   & \color{c1}+14.7  & \color{c1}+10.0   & \color{c1}+14.5  & \color{c1}+6.9      & \color{c1}+14.4     & \color{c1}+6.6      & \color{c1}+12.1        \\ 
                        & Set-CLIP     & \color{c1}+2.1    & \color{c1}+2.9   & \color{c1}+1.9    & \color{c1}+3.6   & \color{c1}+4.3      & \color{c1}+12.3     & \color{c1}+5.1      & \color{c1}+12.6        \\ \bottomrule
\end{tabular}}
\captionof{table}{Benchmark results on general vision-language field. CLIP(1/3) is the baseline and values highlighted in \textcolor{c1}{green} indicate the improvements. Set-CLIP brings benefits across diverse settings, especially in Mini COCO dataset with VIT as encoder.}
\label{table5}
\vspace{-0.3cm}
\end{minipage}%
\hspace{7pt}
\begin{minipage}{.3\textwidth}
\vspace{-0.25cm}
\centering
\raisebox{0.0cm}{\includegraphics[width=1\textwidth]{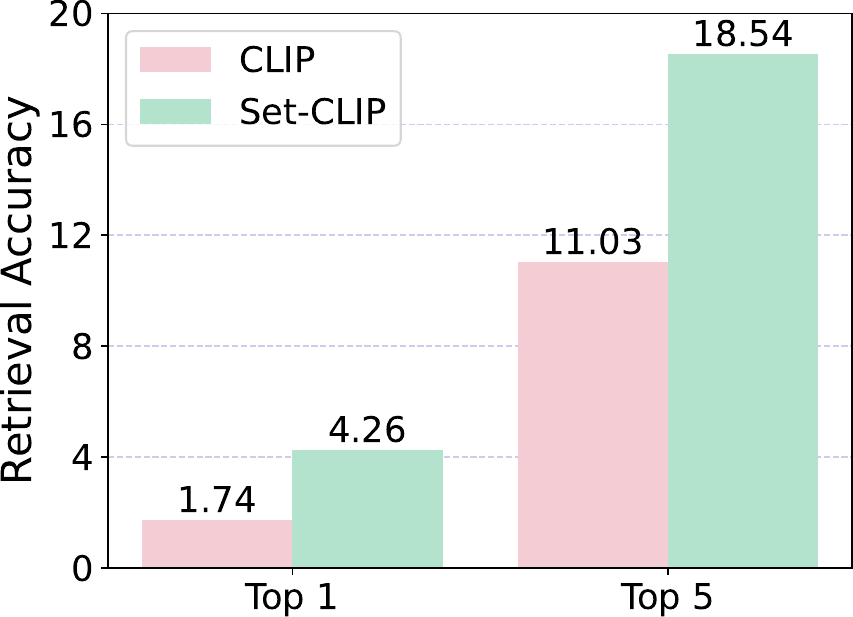}}
\captionof{figure}{Retrieval results of CLIP and Set-CLIP under completely unsupervised scenario.}
\label{fig3}
\vspace{-0.3cm}
\end{minipage}
\end{figure*}

\textbf{Results:}
Table \ref{table3} displays the results of zero-shot classification and the first five lines are from \cite{s-clip}. In this experiment, Set-CLIP is designed based on S-CLIP and trained to narrow the embedding distribution gap between unlabeled images and texts under the guidance of SDD. The whole distributions of batches from different modalities may keep similar even though the explicit matching relationship is unknown between specific samples. For zero-shot classification, our method shows outstanding performance except RSSCN7. We can also find that existing methods are all difficult to bring much gain compared with other datasets, so it is believed that RSSCN7 may have significant gap with training set resulting in greater difficulty for inference. As shown in Appendix G, Set-CLIP consistently improves the results in image-text retrieval, which proves that less distribution gap between unlabeled images and texts brings robust pseudo-labels and stable distribution structure.   



\subsection{Evaluation On General Vision-Language Retrieval}
\textbf{Overview of tasks and training setup:}
Besides above specialized domains, we also evaluate the performance of Set-CLIP in general vision-language field. Experiments are carried on Flickr-8k\cite{flickr-8k} and Mini COCO while image-text retrieval is adopted as downstream task. Moreover, the vision encoder employs ResNet-50 or ViT-32\cite{vit} and BERT\cite{bert} acts as the text encoder. We choose the first description of each image as corresponding caption while dropout ratio is set to $0.3$ for augmentation following \cite{simcse}. Pretraining process lasts $50$ epochs and batch size is $64$ with learning rate equalling to $0.001$. Furthermore, $1/3$ multimodal data is considered to be matched while the rest data is unlabeled. Datailed statements of datasets will be displayed in Appendix C.


\textbf{Results:}
We evaluate the performances with different models as well as datasets and the results is shown in Table \ref{table5}. CLIP(1/3) is trained by 1/3 dataset while CLIP(1) learns alignment on the whole dataset. Set-CLIP employs 1/3 matched data for supervised learning and enlarge knowledge range from rest unpaired data. For better observing effects of different strategies, we adopt the improvement value as yardstick and baseline is CLIP(1/3). We can find that Set-CLIP continuously brings gains regardless of settings and the overall performance of VIT is better than ResNet for given datasets. From the experimental results, we may also conclude that multimodal fusion could be simple when different modality models have similar distance measurement. 

\begin{table}[]
\centering
\scalebox{0.95}{
\begin{tabular}{cccccc}
\toprule
\multirow{2}{*}{SSL} & \multirow{2}{*}{SDD} & \multicolumn{3}{c}{Fold Classification} & \multirow{2}{*}{\begin{tabular}[c]{@{}c@{}}EC\end{tabular}} \\ \cline{3-5}
                             &                      & Fold     & Super     & Family     &                                                                              \\ \midrule
\scalebox{0.75}{\usym{2613}}   & \scalebox{0.75}{\usym{2613}}                     & 57.7         & 78.6                & 99.6           & 0.881                                                                             \\
 \checkmark                     & \scalebox{0.75}{\usym{2613}}                     & 57.9         & 78.7                & 99.6           & 0.881                                                                             \\
\scalebox{0.75}{\usym{2613}}   & \checkmark                     & 58.5         & \textbf{80.1}                & 99.6           & 0.878                                                                             \\
\checkmark                     & \checkmark                     & \textbf{59.1}         & 79.7                & \textbf{99.6}           & \textbf{0.884}                                                                             \\ \bottomrule
\end{tabular}}
\caption{Ablation results evaluated on protein representation field to analyze the roles of SSL and SDD. Thereinto, Super denotes Superfamily task and EC is EC number prediction. The values in \textbf{bold} is the best result at each task.}
\label{table6}
\vspace{-0.5cm}
\end{table}


\subsection{Ablation}
Set-CLIP applies different kinds of objectives to explore implicit semantic alignment from low-aligned multimodal data and shows excellent performances in various experiments. Then, we will further analyze the internal mechanism by replying to the following noteworthy questions. 

\textbf{Q1: How will SDD and SSL influence the final effectiveness?} To answer this question, we conduct experiments in protein representation field and results are shown in Table~\ref{table6}. We can find that Set-CLIP trained with both objectives shows better performance except superfamily classification while the model only trained by SDD outperforms at this item but fall into negative optimization in EC number prediction. Single SSL brings weak improvement compared with the baseline so it also acquires additional alignment information during pretraining process. Through above results, we believe that the combination of these two objectives brings more advantages rather than simple stack of respective effect, in other words,  these two losses interact and depend on each other. Specifically, SDD can exploit alignment information in unsupervised data but may cause mode collapse\cite{zhaohang} and negative learning as shown at the third line. While adding SSL can further constrain the optimization direction and increase the stability of overall distribution. However, SSL will also affect the latent distribution of similar samples, hence it is necessary to balance the relationship to achieve better performance\cite{plato}.

\begin{table}[]
\centering
\scalebox{0.95}{
\begin{tabular}{ccccccc}
\toprule                                                              
\multirow{2}{*}{RD} & \multirow{2}{*}{KL} & \multicolumn{2}{c}{I$\rightarrow$T R@3} & \multicolumn{2}{c}{T$\rightarrow$I R@3} \\ \cline{3-6} 
&                     & 50                     & 100                    & 50                     & 100                    \\ \midrule
\scalebox{0.75}{\usym{2613}}                                                                               & \scalebox{0.75}{\usym{2613}}                   & 28.4                   & 18.4                   & 29.6                   & 18.5                   \\
\scalebox{0.75}{\usym{2613}}                                                                               & \checkmark                   & 29.1                   & 18.7                   & 30.6                   & 18.6                   \\
\checkmark                                                                               & \scalebox{0.75}{\usym{2613}}                   & 28.7                   & 18.8                   & 29.9                   & 18.9                   \\
\checkmark                                                                               & \checkmark                   & \bf{29.8}                   & \bf{20.2}                   & \bf{31.1}                   & \bf{20.3}                   \\ \bottomrule
\end{tabular}}
\caption{Ablation results on Flickr-8k with ResNet as image encoder to analyze the effects of key modules in SDD. \textbf{Bold} indicates the best result. RD denotes relative distance while $50$ and $100$ represent the retrieval size.}
\label{table8}
\vspace{-0.5cm}
\end{table}

\textbf{Q2: How is the performance if we change some modules of SDD?}
With fine-grained distribution similarity measurement, SDD plays a critical role in semi-supervised fusion. So it is essential to deconstruct SDD and analyse which settings may lead to better effects. We make retrieval experiment on Flickr-8k with Top-3 recall and Table~\ref{table8} display the results. Relative distance(RD) $\| x-t_i \|^2/\sigma(T)$ in eq.~\ref{five} can eliminate the indistinguishability in tight latent cluster compared to absolute distance while Kullback-Leibler Divergence(KL) in eq.~\ref{four} may be more suitable for distribution contrast with MSE. It is clear that models achieve the greatest performance when adding RD and KL simultaneously. Figure~\ref{fig3} shows retrieval results of CLIP and Set-CLIP with no paired data for supervised training and it is obvious our designed objectives can still guide to explore alignment in pairing scarcity scenario, especially bring improvement of $144.83\%$ with Top-1 recall. Other ablation results will be shown in Appendix I.  


\section{Conclusion}
We reframe semi-supervised multimodal alignment as manifold matching issue and propose a new method named as Set-CLIP. Based on the data itself, we design novel pretraining tasks to explore latent alignment from unpaired multimodal data in a fine-grained manner. Through extensive experiments across various fields, we demonstrate the superiority of our method to realize rubost generalization, which provides a possible way for multimodal fusion in specialized domains with insufficient aligned data.

\bibliography{aaai25}

\section{Reproducibility Checklist}
\textbf{This paper:}
\begin{itemize}
    \item Includes a conceptual outline and/or pseudocode description of AI methods introduced (yes) 
    \item Clearly delineates statements that are opinions, hypothesis, and speculation from objective fact and results (yes)
    \item Provides well marked pedagogical references for less-familiar readers to gain background necessary to replicate the paper (yes)
\end{itemize}
\textbf{Does this paper make theoretical contributions?} (yes)\\
If yes, please complete the list below.
\begin{itemize}
    \item All assumptions and restrictions are stated clearly and formally. (yes)
    \item All novel claims are stated formally (e.g., in theorem statements). (yes)
    \item Proofs of all novel claims are included. (yes)
    \item Proof sketches or intuitions are given for complex and/or novel results. (yes)
    \item Appropriate citations to theoretical tools used are given. (yes)
    \item All theoretical claims are demonstrated empirically to hold. (yes)
    \item All experimental code used to eliminate or disprove claims is included. (yes)
\end{itemize}
\textbf{Does this paper rely on one or more datasets?} (yes)\\
If yes, please complete the list below.
\begin{itemize}
    \item A motivation is given for why the experiments are conducted on the selected datasets (yes)
    \item All novel datasets introduced in this paper are included in a data appendix. (yes)
    \item All novel datasets introduced in this paper will be made publicly available upon publication of the paper with a license that allows free usage for research purposes. (yes)
    \item All datasets drawn from the existing literature (potentially including authors’ own previously published work) are accompanied by appropriate citations. (yes)
    \item All datasets drawn from the existing literature (potentially including authors’ own previously published work) are publicly available. (yes)
    \item All datasets that are not publicly available are described in detail, with explanation why publicly available alternatives are not scientifically satisfying. (yes)
\end{itemize}
\textbf{Does this paper include computational experiments?} (yes)\\
If yes, please complete the list below.
\begin{itemize}
    \item Any code required for pre-processing data is included in the appendix. (yes)
    \item All source code required for conducting and analyzing the experiments is included in a code appendix. (yes)
    \item All source code implementing new methods have comments detailing the implementation, with references to the paper where each step comes from (yes)
    \item If an algorithm depends on randomness, then the method used for setting seeds is described in a way sufficient to allow replication of results. (yes)
    \item This paper specifies the computing infrastructure used for running experiments (hardware and software), including GPU/CPU models; amount of memory; operating system; names and versions of relevant software libraries and frameworks. (yes)
    \item This paper formally describes evaluation metrics used and explains the motivation for choosing these metrics. (yes)
    \item This paper states the number of algorithm runs used to compute each reported result. (yes)
    \item Analysis of experiments goes beyond single-dimensional summaries of performance (e.g., average, median) to include measures of variation, confidence, or other distributional information. (yes)
    \item The significance of any improvement or decrease in performance is judged using appropriate statistical tests (e.g., Wilcoxon signed-rank). (yes)
    \item This paper lists all final hyper-parameters used for each model/algorithm in the paper’s experiments. (yes)
    \item This paper states the number and range of values tried per (hyper-) parameter during development of the paper, along with the criterion used for selecting the final parameter setting. (yes)
\end{itemize}
\end{document}